\documentclass{article}
\usepackage{kotex}
\usepackage{spconf,amsmath,graphicx}
\usepackage{multirow}
\usepackage[verbose]{microtype}

\title{A DECODING SCHEME WITH SUCCESSIVE AGGREGATION OF  \\MULTI-LEVEL features FOR LIGHT-WEIGHT SEMANTIC SEGMENTATION}
\name{Jiwon Yoo \qquad Jangwon Lee \qquad Gyeonghwan Kim$^{\dagger}$\thanks{$^{\dagger}$Corresponding author.}}
\address{Department of Electronic Engineering, Sogang University, Seoul 04107, Republic of Korea}
\begin{document}
%
\maketitle
\begin{abstract}
Multi-scale architecture, including hierarchical vision transformer, has been commonly applied to high-resolution semantic segmentation to deal with computational complexity with minimum performance loss. 
In this paper, we propose a novel decoding scheme for semantic segmentation in this regard, which takes multi-level features from the encoder with multi-scale architecture. 
The decoding scheme based on a multi-level vision transformer aims to achieve not only reduced computational expense but also higher segmentation accuracy, by introducing successive cross-attention in aggregation of the multi-level features.
Furthermore, a way to enhance the multi-level features by the aggregated semantics is proposed. 
The effort is focused on maintaining the contextual consistency from the perspective of attention allocation and brings improved performance with significantly lower computational cost.
Set of experiments on popular datasets demonstrates superiority of the proposed scheme to the state-of-the-art semantic segmentation models in terms of computational cost without loss of accuracy, and extensive ablation studies prove the effectiveness of ideas proposed.

\end{abstract}

\section{Introduction}
\label{sec:intro}

Semantic segmentation is a fundamental task in computer vision that predicts the category of each pixel in an image, which has been utilized in various fields such as autonomous driving \cite{autonomous_driving} and medical image analysis\cite{medical_image_analysis}. 
As the vision Transformer (ViT)\cite{ViT} has shown notable achievements in the entire field of computer vision, application of ViT to semantic segmentation can be frequently observed as well. 
However, single-scale feature representation and multi-head self-attention (MHA), which inherently reside in ViT, lead to high computational cost for large images and it hinders practical implementation of the semantic segmentation.

\begin{figure}[t]
\includegraphics[width=\columnwidth]{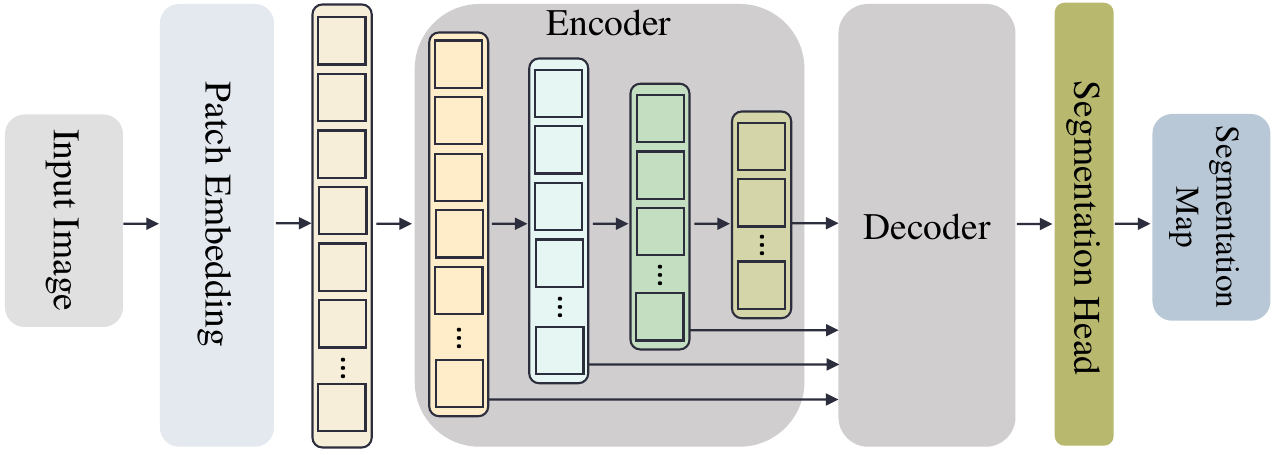}
\caption{
The structure of semantic segmentation scheme that employs HVT-based encoder. The performance relies on how the decoder effectively  fuses the features of different layers.}
\label{semsegstruct} 
\end{figure}
Models based on hierarchical vision Transformer (HVT)\cite{PVT,Swin,LVT,PoolFormer}, in which attention mechanisms with multi-scale features are involved,  have been proposed to lower the computational complexity of the original attention scheme and hence resulting in the prevalent adoption of HVT models as encoders for semantic segmentation as well. 
Fig.~\ref{semsegstruct} shows the structure of semantic segmentation that employs HVT-based encoder.
The performance of those semantic segmentation approaches, therefore, strongly relies on how the decoder focuses on exploring the multi-scale features for the segmentation in effective manner.  
So far, the vast number of methods devised have primarily focused on that aspect.

\begin{figure*}[t]
\includegraphics[width=\textwidth]{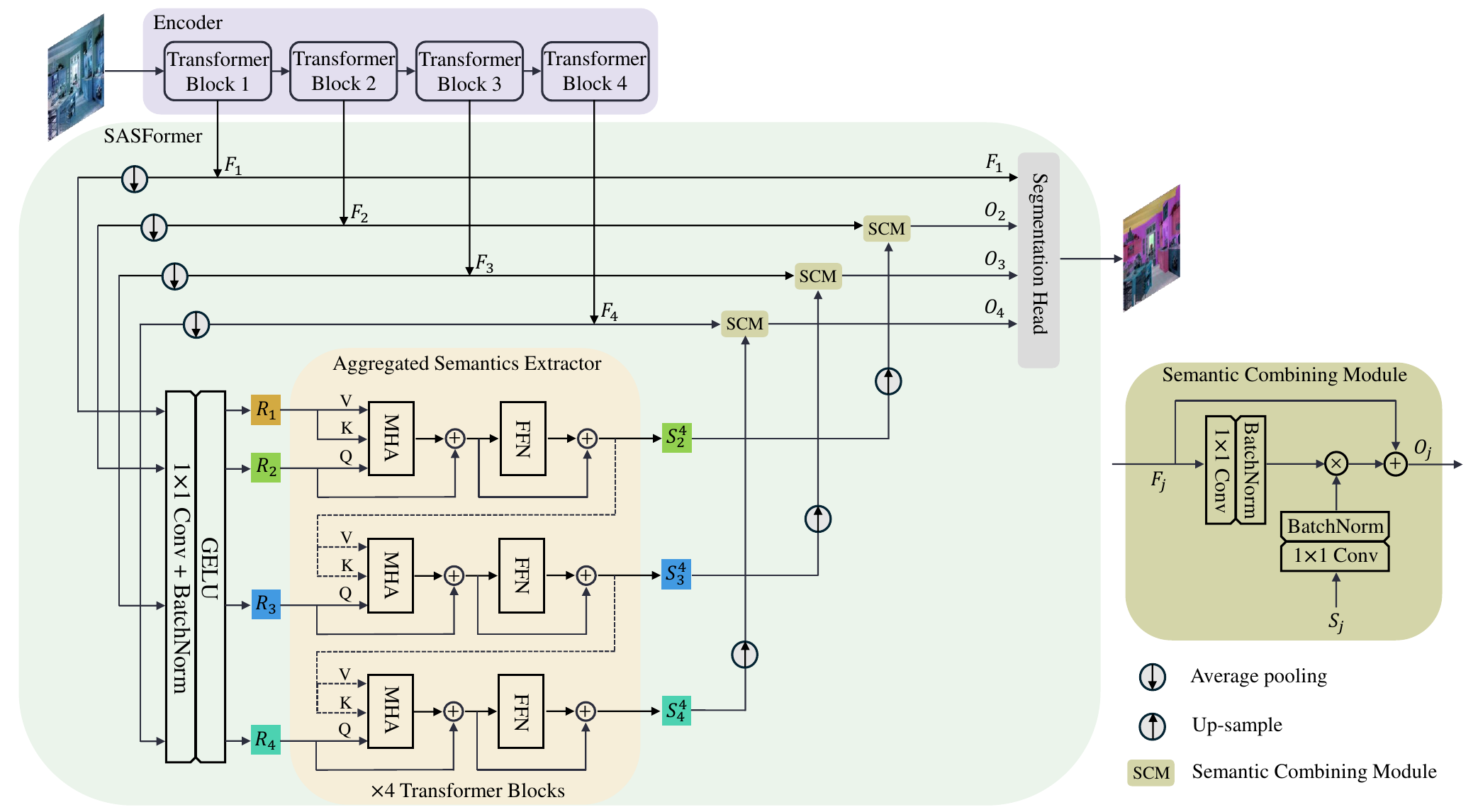}
\caption{The overall structure of the HVT-based semantic segmentation where the proposed SASFormer is used as the decoder. Q, K, and V represent query, key, and value, respectively. 
The figure illustrates how the successive aggregation of the multi-scale features is performed and the aggregated semantics are employed in the proposed ASE and SCM, respectively.}
\label{proposedstructure} 
\end{figure*}

SegFormer\cite{Segformer} introduces an efficient self-attention in an HVT-based encoder, which is called Mix Transformer (MiT),  with a lightweight all multilayer perceptron (ALL-MLP) decoder. 
It suggests a way of fusing multi-level features through the simple MLP decoder thanks to the large receptive field of the features generated in the encoder, but the improvement is quite limited. 
The lightweight ALL-MLP decoder potentially leads to lower efficiency as it relies solely on increasing the model capacity of the encoder to improve performance. 
Therefore, models with more optimized decoders have been proposed. 
For example, the Lawin Transformer\cite{Lawin} is designed with an optimized decoder using Large window (Lawin) attention. 
Lawin attention captures supplemented multi-scale contextual information through various window sizes, which leads to reduced computational resources. 
In the implementation, the features from different levels are concatenated in the early stage of the decoder.
However, the ways of concatenating and handling the results can be inefficient in terms of maintaining contextual consistency in the attention allocation. 
Furthermore, for large size images, the number of channels is increased due to the increased size of the aggregated feature map, leading to a limitation in reducing computational complexity. 

In this paper, therefore, we propose a novel decoding scheme for semantic segmentation that aims to achieve not only less computational expense but also higher segmentation accuracy.
Fig.~\ref{proposedstructure} shows the overall structure of the proposed scheme: {\bf\em SASFormer} - {\bf S}uccessive {\bf A}ggregation of multi-level features for semantic {\bf S}egmentation based on a hierarchical vision Trans{\bf Former}.
The aggregated semantics extractor (ASE) in the figure plays a central role in SASFormer:
1) performing successive cross-attention to extract aggregated semantics which maintain contextual consistency in terms of the attention allocation for improving segmentation accuracy, and 
2) active utilization of downsampled feature maps from different levels for reducing computational cost significantly.
On the other hand, the semantic combining module (SCM) is devised to utilize the aggregated semantics as weights for enhancing the multi-scale features, hence to improve the accuracy.
The key contributions of the proposed scheme are summarized in the following:


\begin{itemize}
\item A novel decoding scheme for semantic segmentation based on a hierarchical vision Transformer - SASFormer.
\item Successive aggregation of multi-scale features to achieve both improving segmentation accuracy and reducing computational cost by introducing the aggregated semantics extractor (ASE) and the semantic combining module (SCM).
\item  Demonstrating an excellent balance between accuracy and computational expense by comparison with the state-of-the-art (SOTA) models on the ADE20K and Cityscapes datasets, and the effectiveness of the introduced concepts as demonstrated by extensive ablation studies.


\end{itemize}

\section{Related Work}
\label{sec:related work}

\subsection{Hierarchical Vision Transformer}
\label{ssec:HVT}

ViT\cite{ViT}  is the first vision transformer model that achieves great performance in image classification. However, direct application of ViT is not suitable for semantic segmentation tasks that require high computation, as it generates feature maps from a fixed resolution. To address these issues, HVT-based models have been proposed. These models not only generate multi-scale features but also propose efficient attention methods that reduce computational cost. PVT\cite{PVT} and Swin\cite{Swin} use patch merging to generate multi-scale features. They respectively introduce spatial-reduction attention and shifted window to make self-attention more computationally efficient for high-resolution images. 
Subsequently, models like LVT\cite{LVT} and PoolFormer\cite{PoolFormer} aim to improve attention efficiency while maintaining similar architectural structures. Nonetheless, a notable limitation of HVT models stems from the lack of acquiring contextual information from the multi-level features, as they only conduct self-attention on the features inputted at each stage.

\subsection{Semantic segmentation}
\label{ssec:semantic segmentation}

Semantic segmentation involves classifying each pixel in an image into a specific object category. 
As HVT models have been introduced, they are adopted as encoders for semantic segmentation. 
This has led to the development of models equipped with decoders specifically designed to improve multi-level connections. 
The SegFormer\cite{Segformer}, for instance, attempts to improve these connections with an All-MLP decoder, but the overall performance is limited by the characteristics of the MLP.  
Models like SenFormer\cite{SenFormer}, MaskFormer\cite{maskformer}, and Mask2Former\cite{mask2former}, which have been proposed after SegFormer\cite{Segformer}, introduce transformer-based decoders to refine multi-scale features through learnable query sets. 
However, they result in significant computational costs due to an extra pixel decoder in addition to the transformer-based decoder. 
The Lawin Transformer\cite{Lawin}, which employs a transformer-based decoder, introduces large window attention (Lawin attention) to capture contextual information. 
This approach brings improved performance and reduced computational cost, but there seems to be room for improvement in terms of dealing with the high-resolution features which are regarded as the major burden.
Therefore, we aim to design a transformer-based decoder that enhances attention mechanism with multi-level features to improve performance and reduce computational cost. 

\section{THE PROPOSED METHOD}
\label{sec:method}

This section contains the details on the proposed model. 
Descriptions on the two key modules that constitute the model, the aggregated semantics extractor (ASE) and the semantic combining module (SCM), are followed by the overall structure. 

\subsection{Overall Architecture}
\label{ssec:architecture}

The proposed model for light-weight semantic segmentation is shown in Fig.~\ref{proposedstructure}.
The encoder deals with an input image size of $H\times W\times 3$. 
The four-stage encoder generates multi-scale features, $F_{i}$ size of $\frac{H}{2^{i+1}} \times \frac{W}{2^{i+1}} \times C_{i}$, where  $i \in \{1, \ldots, 4\}$ represents the stage index. 
In SASFormer, each of the multi-scale features is resized to a fixed size before passing to ASE.
ASE extracts aggregated semantics from the resized features.  
Subsequently, the aggregated semantics are combined with the corresponding multi-scale features through SCM, and then used to form the final segmentation map through the segmentation head.

\begin{table*}[t]
\caption{Performance comparison of the proposed scheme on ADE20K-val. and Cityscapes-val. with the state-of-the-art models. Boldfaced numbers, which indicate the best performance, demonstrate the overall superiority of SASFormer in terms of the computational efficiency and the accuracy.}
\centering
{\small
\begin{tabular}{|c|lcccccc|}
\hline
\multicolumn{1}{|l|}{}               & \multicolumn{1}{c|}{\multirow{2}{*}{Method}} & \multicolumn{1}{c|}{\multirow{2}{*}{Encoder}} & \multicolumn{1}{c|}{\multirow{2}{*}{Params (M)}} & \multicolumn{2}{c|}{ADE20K}                             & \multicolumn{2}{c|}{Cityscapes}             \\ \cline{5-8} 
\multicolumn{1}{|l|}{}               & \multicolumn{1}{c|}{}                        & \multicolumn{1}{c|}{}                         & \multicolumn{1}{c|}{}                            & \multicolumn{1}{c|}{GFLOPs} & \multicolumn{1}{c|}{mIoU} & \multicolumn{1}{c|}{GFLOPs} & mIoU          \\ \hline
\multirow{9}{*}{\rotatebox[origin=c]{90}{Light-weight config.}} & DeepLabV3+~\cite{deeplabv3}                        & MobileNetV2                                   & 15.4                                             & 69.4                        & 34.0                      & 555.4                       & 75.2          \\
                                     & SwiftNetRN~\cite{swiftnetrn}                       & ResNet-18                                     & 11.8                                             & -                           & -                         & 104.0                       & 75.4          \\
                                     & SegFormer~\cite{Segformer}                      & MiT-B0                                        & 3.8                                              & 8.4                         & 37.4                      & 125.5                       & 76.2          \\
                                     & RTFormer~\cite{rtformer}                              & RTFormer-Slim                                 & 4.8                                              & 17.5                        & 36.7                      & -                           & 76.3          \\
                                     & SegFormer-HILA~\cite{hial}                        & MiT-B0                                        & 4.2                                              & 9.8                         & 38.3                      & 139.4                       & 77.2          \\
                                     & Lawin Transformer~\cite{Lawin}                     & MiT-B0                                        & 4.1                                              & 5.3                         & 38.9                      & 99.3                        & 77.2          \\
                                     & SegFormer-LVT~\cite{LVT}                        & LVT                                           & 3.9                                              & 10.6                        & 39.3                      & 140.9                       & 77.6          \\
                                     & IFA~\cite{IFA}                                   & ResNet-50                                     & 27.8                                             & -                           & -                         & 186.9                       & 78.0          \\
                                     & \textbf{SASFormer(Ours)}                     & \textbf{MiT-B0}                               & \textbf{4.8}                                     & \textbf{5.0}                & \textbf{40.1}             & \textbf{99.3}               & \textbf{78.3} \\ \hline
\multirow{9}{*}{\rotatebox[origin=c]{90}{Middle-weight config.}}    & SenFormer~\cite{SenFormer}                             & Swin-T                                        & 59.0                                             & 179.0                       & 46.0                      & -                           & -             \\
                                     & SegFormer~\cite{Segformer}                             & MiT-B2                                        & 27.5                                             & 62.4                        & 46.5                      & 717.1                       & 81.0          \\
                                     & RTFormer~\cite{rtformer}                              & RTFormer-Base                                 & 16.8                                             & 67.4                        & 42.1                      & -                           & 79.3          \\
                                     & IFA~\cite{IFA}                                   & ResNet-101                                    & 46.7                                             & 67.1                        & 45.2                      & -                           & -             \\
                                     & SegFormer-HILA~\cite{hial}                       & MiT-B2                                        & 30.8                                             & 76.5                        & 46.0                      & 704.2                       & 81.5          \\
                                     & MaskFormer~\cite{maskformer}                           & Swin-T                                        & 42.0                                             & 55.0                        & 46.7                      & -                           & -             \\
                                     & Maks2Former~\cite{mask2former}                          & Swin-T                                        & 47.0                                             & 74.0                        & 47.7                      & -                           & -             \\
                                     & Lawin Transformer~\cite{Lawin}                     & MiT-B2                                        & 29.7                                             & 45.0                        & 47.8                      & 562.8                       & 81.7          \\
                                     & \textbf{SASFormer(Ours)}                     & \textbf{MiT-B2}                               & \textbf{30.7}                                    & \textbf{36.6}               & \textbf{48.0}                & \textbf{512.7}              & \textbf{81.7}    \\ \hline
\end{tabular}%
}
\label{SOTA} 
\end{table*}

\subsection{Accumulated Semantics Extractor (ASE)}
\label{ssec:ASE}
The ASE is carefully designed to optimize the trade-off between computational burden and efficient utilization of the features from different levels. 
Direct utilization of the multi-scale features can lead to higher segmentation accuracy, but it causes higher computational complexity due to the large dimension of the features.
To mitigate the high computational burden, the multi-scale features $F_{i}$, $i \in \{1, \ldots, 4\}$, from the four-stage decoder are downsampled to the single-scale features, $R_{i}$,  with a fixed size of $\frac{H}{64} \times \frac{W}{64} \times C_{i}$, where $C_{i}$ represents the number of channels in the level. 
It is the successive cross-attention that we devise not only to compensate for the information loss occurring during downsampling but also to maintain the contextual consistency among the features from different levels.

The ASE consists of four transformer blocks, each of which contains MHA-based successive cross-attention (SCA) and feed-forward network (FFN), as shown in Fig.~\ref{proposedstructure}.
The MHA layer with Q(query), K(key), V(value), which are generated by linear projection of the features, is processed as follows:
\begin{equation}
	\text{MHA}(Q, K, V) = \text{Softmax}\left(\frac{QK^T}{\sqrt{d_{k}}}\right)V    
\end{equation}
where, $d_{k}$ denotes the embedding dimension of the keys.
%
%
%
The MHA layer and the FFN are connected as shown in Fig.~\ref{proposedstructure}, and the operation can be described by (\ref{eq:sca1}) and (\ref{eq:sca2}):
\begin{flalign}
	A^l_{j} &= \text{MHA}(\text{LN}(KV^l_{j}),\text{LN}(Q^l_{j})) + Q^l_{j} \label{eq:sca1}\\ 
	S^l_{j} &= \text{FFN}(\text{LN}(A^l_{j})) + A^l_{j} \label{eq:sca2}
\end{flalign} 
where $l \in \{1, \ldots, 4\}$ represents the transformer block index, $j \in \{1, 2, 3\}$ represents the index of the feature used as a query, and LN denotes layer normalization. 
For the first transformer block, that is $l=1$,  $KV^l_{j}$ and $Q^l_{j}$ are represented as (\ref{eq:sca3}),
\begin{align}
Q_{j} = R_{j+1},\;
KV_{j} = \begin{cases}
R_{j}, & \text{if } j = 1 \\
S_{j}, & \text{otherwise}
\end{cases} \label{eq:sca3}
\end{align} 
and for the remaining transformer blocks, $KV^l_{j}$ and $Q^l_{j}$ are represented as (\ref{eq:sca4}).
\begin{align}
Q^l_{j} = S^l_{j+1},\;
KV^l_{j} = \begin{cases}
R_{j}, & \text{if } j = 1 \\
S^l_{j}, & \text{otherwise}
\end{cases} \label{eq:sca4}
\end{align}

In the FFN, we enhance the local connections by integrating a depth-wise convolution layer between the two $1 \times 1$ convolution layers such as \cite{shuffle}. The channel expansion ratio  is set to 4.

It needs to be noted that the process represented by (\ref{eq:sca1}) through (\ref{eq:sca4}) is the successive cross-attention (SCA), which is different from a self-attention and a usual cross-attention mechanisms, in terms of how to set the query, key and value.  
In this regard, we can term \{{$S^4_{2}$, $S^4_{3}$, $S^4_{4}$}\} as the aggregated semantics, which are extracted through the four transformer blocks. The successive process over different levels effectively aggregates semantics enriched with contextual information.  


\subsection{Semantic Combining Module (SCM)}
\label{ssec:FEM}

The SCM is introduced to refine the contextual information of the multi-scale features. 
As shown in the right side of Fig.~\ref{proposedstructure}, for $j \in \{2, 3, 4\}$, SCM combines the aggregated semantics $S^4_{j}$, extracted from the ASE with the corresponding multi-scale features $F_{j}$ to enhance the representation.
During this process, the aggregated semantics act as weights and are multiplied with the corresponding multi-scale features, except for the lowest-level feature $F_{1}$.


To treat the aggregated semantics as the weights, they need to be upsampled, by the factor of the downsampling described in \ref{ssec:ASE}, to match the dimension of the corresponding multi-scale features. 
Meanwhile, $F_{j}$ and $S^4_{j}$ are passed through the $1 \times 1$ convolution layer, followed by a batch normalization as illustrated in the figure. 
The process of obtaining the enhanced features $O_{j}$ can be described as (\ref{eq:scm}):
\begin{equation}
	O_j = F_j \times S_j + F_j \text{,~~where~} j=2,3,4   \label{eq:scm} 
\end{equation}
In (\ref{eq:scm}), adding $F_{j}$ to the multiplied result is to make up the loss incurred during the downsampling $F_{j}$ to $R_{j}$.
The enhanced features are concatenated along with $F_{1}$ and through the segmentation head to predict the segmentation map. 


\section{EXPERIMENTS}
\label{sec:pagestyle}

\subsection{Datasets}
\label{ssec:dataset}

Two popular datasets are used for the extensive experiment and the ablation study: ADE20K\cite{ADE20K} and Cityscapes\cite{cityscapes}. The ADE20K dataset consists of 150 categories, with a total of 25K images divided into 20K for training, 2K for validation, and 3K for testing. The Cityscapes dataset consists of 19 categories, with a total of 5,000 images divided into 2975 for training, 500 for validation, and 1525 for testing.

\subsection{Implementation Details}
\label{ssec:detail}
The performance evaluation and the ablation study have been performed based on MMsegmentation\cite{MMSegmentation_Contributors_OpenMMLab_Semantic_Segmentation_2020} with two RTX 3090 GPUs.  
We used MiT\cite{Segformer}, pre-trained on the Imagenet-1K dataset, as the encoder. 
For all datasets and the evaluation, training was conducted over 160K iterations using the AdamW optimizer. Instead of using the BatchNorm layer, we adopted Synchronized BatchNorm, which enables the aggregation of the mean and standard deviation of BatchNorm across multiple GPUs during training. The batch size was set to 16 for the ADE20K dataset and 8 for the Cityscapes dataset. The initial learning rate was set to 0.0001, and a poly LR schedule with a factor of 1.0 was used. The data augmentation method used in SegFormer was applied. The mIoU and FLOPs were reported with single-scale inference, and the performance of the proposed model was evaluated using  ADE20K and Cityscapes validation datasets.

\begin{table}[t]
\caption{Ablation study on the effectiveness of the {\em successive} cross-attention to self- and cross-attentions.}
\centering
{%
\begin{tabular}{lllcllclllcll}
\hline
\multicolumn{3}{c}{Method}                      & \multicolumn{3}{c}{Parmas(M)}   & \multicolumn{4}{c}{GFLOPs}  & \multicolumn{3}{c}{mIoU}  \\ \hline
\multicolumn{3}{l}{Self-attention}                    & \multicolumn{3}{c}{7.2}          & \multicolumn{4}{c}{5.5}    & \multicolumn{3}{c}{40.0}       \\
\multicolumn{3}{l}{Cross-attention} & \multicolumn{3}{c}{4.8}          & \multicolumn{4}{c}{5.0}    & \multicolumn{3}{c}{39.9}       \\
\multicolumn{3}{l}{Successive CA}                        & \multicolumn{3}{c}{4.8} & \multicolumn{4}{c}{5.0}    & \multicolumn{3}{c}{40.1}       \\ \hline
\end{tabular}%
}
\label{ablationsca} 
\end{table}
\begin{table}[t]
\caption{Ablation study on the way of combining in SCM.}
\centering
{%
\begin{tabular}{lllcllclllcll}
\hline
\multicolumn{3}{c}{Method}                      & \multicolumn{3}{c}{Parmas(M)}   & \multicolumn{4}{c}{GFLOPs}  & \multicolumn{3}{c}{mIoU}  \\ \hline
\multicolumn{3}{l}{Eq. (\ref{eq:scmalt1})}                    & \multicolumn{3}{c}{4.8}          & \multicolumn{4}{c}{5.0}    & \multicolumn{3}{c}{39.5}       \\
\multicolumn{3}{l}{Eq. (\ref{eq:scmalt2})}  & \multicolumn{3}{c}{4.8}          & \multicolumn{4}{c}{5.0}    & \multicolumn{3}{c}{39.6}       \\
\multicolumn{3}{l}{Eq. (\ref{eq:scm})}  & \multicolumn{3}{c}{4.8} & \multicolumn{4}{c}{5.0}    & \multicolumn{3}{c}{40.1}       \\ \hline
\end{tabular}%
}
\label{ablationscm} 
\end{table}

\subsection{Comparisons with the State-of-the-Art}
\label{ssec:comparison} 

The performance comparison of the proposed method with the state-of-the-art models on ADE20K and Cityscapes datasets has been carried out.
As shown in Table~\ref{SOTA}, in light-weight configuration, our model achieves 40.1\% mIoU, 5.0 GFLOPs with 4.8M parameters on ADE20K, and 78.3\% mIoU, 99.3 GFLOPs on Cityscapes. 
Compared to SegFormer-B0, SASFormer-B0 shows 2.7\% higher mIoU and 40.4\% less computation on ADE20K, and 1.9\% higher mIoU and 20.9\% less computation on Cityscapes. 
Moreover, compared to the Lawin Transformer-B0, SASFormer-B0 shows 1.2\% higher mIoU with 5.6\% less computation on ADE20K. 
SASFormer-B0 achieves a 1.2\% higher mIoU on Cityscapes, with roughly the same computational cost.

In medium-weight configuration, SASFormer-B2 achieves  48.0\% mIoU and 36.6 GFLOPs with 30.7M parameters on ADE20K, and 81.7\% mIoU and 512.7 GFLOPs on Cityscapes. When compared to SegFormer-B2, SASFormer-B2 shows 41.3\% less computation and 1.5\% higher mIoU on ADE20K, and 28.5\% less computation and 0.9\% higher mIoU on Cityscapes. In comparison with Lawin-Transformer-B2, SASFormer-B2 shows an 18.7\% less computation and 0.2\% higher mIoU on ADE20K. 
SASFormer-B2 achieves the same mIoU as Lawin Transformer-B2, but with 8.9\% less computation on Cityscapes.  
Overall, compared to the SOTA models in semantic segmentation, the carefully designed decoder brings higher mIoU and fewer FLOPs with roughly similar number of parameters.
\begin{figure}[t]
\centering
\includegraphics[width=0.9\columnwidth]{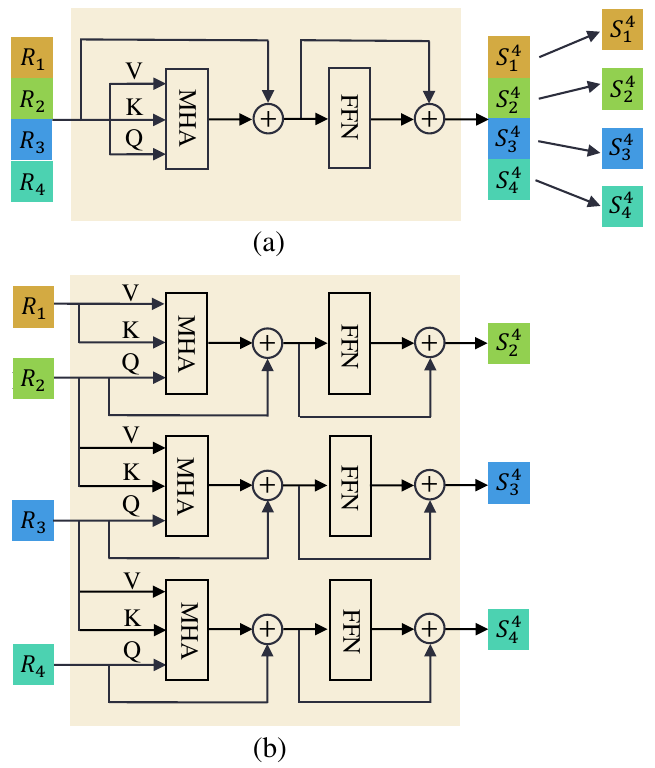}
\caption{Two attention configurations can be adopted in ASE for the ablation study of the successive cross-attention: (a) self-attention, and (b) cross-attention.}
\label{attentionablation} 
\end{figure}

\subsection{Ablation study}
\label{ssec:ablation study}

In the ablation study, we use MiT-B0 as the encoder, and the training and the evaluation are carried out using the corresponding sets of the ADE20K.
For an assessment of the proposed schemes applied to the decoder, all evaluations are conducted under the same conditions as the baseline models.
 

\begin{figure*}[t]
\includegraphics[width=\textwidth]{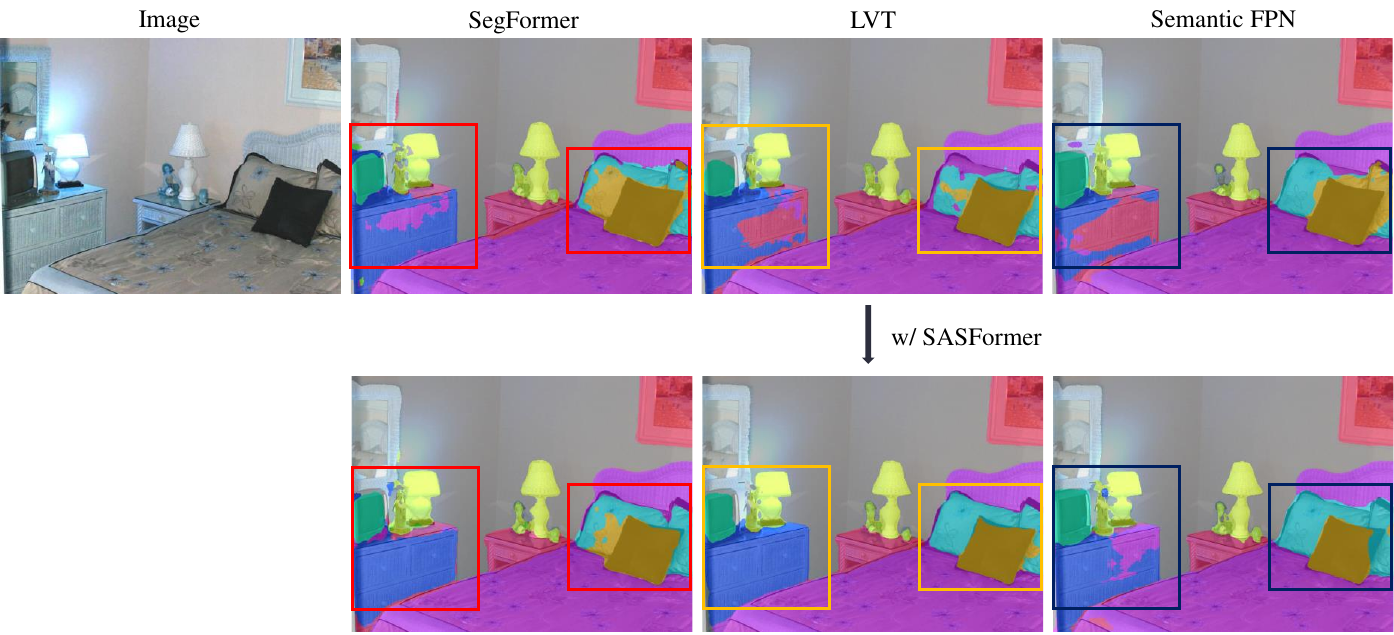}
\caption{Qualitative comparison of an image in ADE20K. The results of three HVT-based models are compared to the ones obtained by the models in which the decoder is replaced with SASFormer. 
For all three, models with the proposed scheme result in more precise segmentation in the boxed areas where multi-scale objects are present.}
\label{qualcomp} 
\end{figure*}

\noindent \textbf{The effectiveness of the {\em successive} cross-attention.} \quad 
To validate the successive manner applied to the cross-attention, experiments with two different attention methods are conducted.
The first one in Fig.~\ref{ablationscm}(a) employs self-attention on the features formed with concatenated $R_i$.
The semantics extracted by the self-attention are split, and all except for the lowest-level one, are provided to the SCM.
The second one employs cross-attention as shown in Fig.~\ref{ablationscm}(b), which lacks the successive connections.
As shown in  Table~\ref{ablationsca}, the SCA results in better performance than both of them. 
Therefore, it demonstrates that the semantics extracted through the proposed successive configuration applied to cross-attention effectively aggregates contextual information by maintaining contextual consistency from the perspective of attention allocation. 


\noindent \textbf{The way of combining in the SCM.} \quad To evaluate the effectiveness of the way of combination implemented as (\ref{eq:scm}), experiments on two different combination configurations, denoted in (\ref{eq:scmalt1}) and (\ref{eq:scmalt2}),  are performed. 
\begin{flalign}
	O_j &= F_j \times S_j \text{,~~where~} j=2,3,4   \label{eq:scmalt1} \\
	O_j &= F_j \times S_j + S_j \text{,~~where~} j=2,3,4   \label{eq:scmalt2} 
\end{flalign} 

As listed in Table~\ref{ablationscm}, the combination scheme in (\ref{eq:scm}) results in higher mIoU than the two configurations, with no change in FLOPs and the number of parameters. 
Therefore, the result proves that the residual connection of the multi-scale features in (\ref{eq:scm}) is effective in minimizing the adverse effect due to the loss incurred during the downsampling.  


\begin{table}[t]
\caption{Ablation study on the number of transformer blocks.}
\centering 
\begin{tabular}{cccc}
\hline
\multicolumn{1}{l}{The no. of blocks} & \multicolumn{1}{l}{Params(M)} & \multicolumn{1}{l}{GFLOPs} & \multicolumn{1}{l}{mIoU} \\ \hline
1                                       & 4.829                           & 4.808                      & 39.2                         \\
2                                       & 4.834                           & 4.881                      & 39.5                         \\
3                                       & 4.839                           & 4.955                      & 39.8                         \\
4                                       & 4.844                           & 5.028                      & 40.1                     \\
5                                       & 4.849                           & 5.101                      & 39.9                         \\ \hline
\end{tabular}%
\label{blocknumber} 
\end{table}

\noindent \textbf{The number of transformer blocks.} \quad 
The performance indices in Table~\ref{blocknumber} are obtained from implementations with a different number of transformer blocks.
Given that the increase in the number of parameters is marginal, the number of blocks is set to 4 as the optimal balance between FLOPs and mIoU.

\noindent \textbf{The applicability of the SASFormer as a decoder.} \quad 
The applicability of an idea is a primary concern in this type of work. To prove the applicability of the proposed decoder to semantic segmentation, a set of experiments has been carried out, where  for several existing HVT-based semantic segmentation models, the  decoders are replaced with SASFormer. 
Results in Table~\ref{Decoder} prove that the proposed decoding scheme is able to accommodate the multi-level features effectively, in terms of not only reduced computational cost but also increased accuracy.
It should be noted that when SASFormer is combined with HVT-based encoders for semantic segmentation, inherent flaws in such models can be handled effectively.   
Fig.~\ref{qualcomp} shows the qualitative comparison of the results of the three HVT-based models to the ones obtained by the models in which the decoder is replaced with SASFormer.   
Boxes in the figure indicate the locations where multi-scale objects are present, and the models equipped with SASFormer demonstrate the capability of precise segmentation in the intricate areas. 

\begin{table}[t]
\caption{Ablation study on the applicability of SASFormer as a decoder.}
\centering \small
\resizebox{\columnwidth}{!}{%
\begin{tabular}{clccc}
\hline
Enocder                         & \multicolumn{1}{c}{Decoder} & Params(M) & GFLOPs & mIoU \\ \hline
\multirow{2}{*}{MiT-B0}         & SegFormer                  & 3.8       & 8.4    & 37.4 \\
                                & SASFormer                  & 4.8       & 37.4   & 40.1 \\ \hline
\multirow{2}{*}{LVT}            & SegFormer                  & 3.9       & 10.6   & 39.3 \\
                                & SASFormer                  & 4.9       & 7.2    & 40.8 \\ \hline
{ PoolFormer} & Semantic FPN~\cite{kirillov2019panoptic}               & 15.7      & 31.0   & 37.2 \\
  -S12~\cite{PoolFormer}                              & SASFormer                  & 17.4      & 26.4   & 39.3 \\ \hline
\end{tabular}%
}
\label{Decoder} 
\end{table}

\section{CONCLUSION}
\label{sec:typestyle}

In this paper, a simple, yet powerful decoder architecture, called SASFormer, for semantic segmentation is proposed. 
The light-weight transformer-based decoder is designed to optimize the trade-off between the computational cost and the segmentation accuracy, by taking into account the way of maintaining contextual consistency from the perspective of attention allocation with successive aggregation of the multi-scale features from the HVT-based encoder.
The effectiveness of the proposed scheme is proved by comparison with various light-weight semantic segmentation models through a set of experiments and extensive ablation studies. 
The authors believe that the proposed scheme is applicable to other computer vision tasks which require the trade-off. 

%


\bibliographystyle{IEEEbib}
\bibliography{refs}

\begin{thebibliography}{10}

\bibitem{autonomous_driving}
Li~Wang, Dong Li, Han Liu, Jinzhang Peng, Lu~Tian, and Yi~Shan,
\newblock ``Cross-dataset collaborative learning for semantic segmentation in autonomous driving,''
\newblock in {\em AAAI}, 2022, vol.~36, pp. 2487--2494.

\bibitem{medical_image_analysis}
Jiahuan Song, Xinjian Chen, Qianlong Zhu, Fei Shi, Dehui Xiang, Zhongyue Chen, Ying Fan, Lingjiao Pan, and Weifang Zhu,
\newblock ``Global and local feature reconstruction for medical image segmentation,''
\newblock {\em IEEE Transactions on Medical Imaging}, vol. 41, no. 9, pp. 2273--2284, 2022.

\bibitem{ViT}
Alexey Dosovitskiy, Lucas Beyer, Alexander Kolesnikov, Dirk Weissenborn, Xiaohua Zhai, Thomas Unterthiner, Mostafa Dehghani, Matthias Minderer, Georg Heigold, Sylvain Gelly, et~al.,
\newblock ``An image is worth 16x16 words: Transformers for image recognition at scale,''
\newblock in {\em ICLR}, 2020.

\bibitem{PVT}
Wenhai Wang, Enze Xie, Xiang Li, Deng-Ping Fan, Kaitao Song, Ding Liang, Tong Lu, Ping Luo, and Ling Shao,
\newblock ``Pyramid vision transformer: A versatile backbone for dense prediction without convolutions,''
\newblock in {\em ICCV}, 2021, pp. 568--578.

\bibitem{Swin}
Ze~Liu, Yutong Lin, Yue Cao, Han Hu, Yixuan Wei, Zheng Zhang, Stephen Lin, and Baining Guo,
\newblock ``Swin transformer: Hierarchical vision transformer using shifted windows,''
\newblock in {\em ICCV}, 2021, pp. 10012--10022.

\bibitem{LVT}
Chenglin Yang, Yilin Wang, Jianming Zhang, He~Zhang, Zijun Wei, Zhe Lin, and Alan Yuille,
\newblock ``Lite vision transformer with enhanced self-attention,''
\newblock in {\em CVPR}, 2022, pp. 11998--12008.

\bibitem{PoolFormer}
Weihao Yu, Mi~Luo, Pan Zhou, Chenyang Si, Yichen Zhou, Xinchao Wang, Jiashi Feng, and Shuicheng Yan,
\newblock ``Metaformer is actually what you need for vision,''
\newblock in {\em CVPR}, 2022, pp. 10819--10829.

\bibitem{Segformer}
Enze Xie, Wenhai Wang, Zhiding Yu, Anima Anandkumar, Jose~M Alvarez, and Ping Luo,
\newblock ``Segformer: Simple and efficient design for semantic segmentation with transformers,''
\newblock {\em NeurIPS}, vol. 34, pp. 12077--12090, 2021.

\bibitem{Lawin}
Haotian Yan, Chuang Zhang, and Ming Wu,
\newblock ``Lawin transformer: Improving semantic segmentation transformer with multi-scale representations via large window attention,''
\newblock {\em arXiv preprint arXiv:2201.01615}, 2022.

\bibitem{SenFormer}
Walid Bousselham, Guillaume Thibault, Lucas Pagano, Archana Machireddy, Joe Gray, Young~Hwan Chang, and Xubo Song,
\newblock ``Efficient self-ensemble for semantic segmentation,''
\newblock {\em arXiv preprint arXiv:2111.13280}, 2021.

\bibitem{maskformer}
Bowen Cheng, Alex Schwing, and Alexander Kirillov,
\newblock ``Per-pixel classification is not all you need for semantic segmentation,''
\newblock {\em NeurIPS}, vol. 34, pp. 17864--17875, 2021.

\bibitem{mask2former}
Bowen Cheng, Ishan Misra, Alexander~G Schwing, Alexander Kirillov, and Rohit Girdhar,
\newblock ``Masked-attention mask transformer for universal image segmentation,''
\newblock in {\em CVPR}, 2022, pp. 1290--1299.

\bibitem{deeplabv3}
Liang-Chieh Chen, Yukun Zhu, George Papandreou, Florian Schroff, and Hartwig Adam,
\newblock ``Encoder-decoder with atrous separable convolution for semantic image segmentation,''
\newblock in {\em ECCV}, 2018, pp. 801--818.

\bibitem{swiftnetrn}
Marin Orsic, Ivan Kreso, Petra Bevandic, and Sinisa Segvic,
\newblock ``In defense of pre-trained imagenet architectures for real-time semantic segmentation of road-driving images,''
\newblock in {\em CVPR}, 2019, pp. 12607--12616.

\bibitem{rtformer}
Jian Wang, Chenhui Gou, Qiman Wu, Haocheng Feng, Junyu Han, Errui Ding, and Jingdong Wang,
\newblock ``R{T}former: Efficient design for real-time semantic segmentation with transformer,''
\newblock {\em NeurIPS}, vol. 35, pp. 7423--7436, 2022.

\bibitem{hial}
Gary Leung, Jun Gao, Xiaohui Zeng, and Sanja Fidler,
\newblock ``Improving semantic segmentation in transformers using hierarchical inter-level attention,''
\newblock {\em arXiv preprint arXiv:2207.02126}, 2022.

\bibitem{IFA}
Hanzhe Hu, Yinbo Chen, Jiarui Xu, Shubhankar Borse, Hong Cai, Fatih Porikli, and Xiaolong Wang,
\newblock ``Learning implicit feature alignment function for semantic segmentation,''
\newblock in {\em ECCV}. Springer, 2022, pp. 487--505.

\bibitem{shuffle}
Zilong Huang, Youcheng Ben, Guozhong Luo, Pei Cheng, Gang Yu, and Bin Fu,
\newblock ``Shuffle transformer: Rethinking spatial shuffle for vision transformer,''
\newblock {\em arXiv preprint arXiv:2106.03650}, 2021.

\bibitem{ADE20K}
Bolei Zhou, Hang Zhao, Xavier Puig, Sanja Fidler, Adela Barriuso, and Antonio Torralba,
\newblock ``Scene parsing through {ADE20k} dataset,''
\newblock in {\em CVPR}, 2017, pp. 633--641.

\bibitem{cityscapes}
Marius Cordts, Mohamed Omran, Sebastian Ramos, Timo Rehfeld, Markus Enzweiler, Rodrigo Benenson, Uwe Franke, Stefan Roth, and Bernt Schiele,
\newblock ``The {Cityscapes} dataset for semantic urban scene understanding,''
\newblock in {\em CVPR}, 2016, pp. 3213--3223.

\bibitem{MMSegmentation_Contributors_OpenMMLab_Semantic_Segmentation_2020}
MMSegmentation Contributors,
\newblock ``Openmmlab semantic segmentation toolbox and benchmark,''
\newblock https://github.com/open-mmlab/mmsegmentation, 2020.

\bibitem{kirillov2019panoptic}
Alexander Kirillov, Ross Girshick, Kaiming He, and Piotr Doll{\'a}r,
\newblock ``Panoptic feature pyramid networks,''
\newblock in {\em CVPR}, 2019, pp. 6399--6408.

\end{thebibliography}

\end{document}